\documentclass[letterpaper, 10 pt, conference]{ieeeconf}  

\IEEEoverridecommandlockouts                              

\UseRawInputEncoding

\usepackage{xspace}
\usepackage{graphicx}
\usepackage{xcolor, pict2e}
\usepackage{soul}
\usepackage{array}
\usepackage{bm}
\usepackage{booktabs}
\usepackage{amsmath}
\usepackage{caption}
\usepackage{subcaption}
\usepackage{cite}
\usepackage{amssymb}
\usepackage{tikz}
\usepackage{stackengine}
\usepackage{transparent}

\newcommand{\tikzcircle}[2][red,fill=red]{\tikz[baseline=-0.5ex]\draw[#1,radius=#2] (0,0) circle ;}

\newcommand{\gist}[1]{}

\newcommand{\mlabel}[1]{\label{#1}}
\newcommand{\mcite}[1]{\cite{#1}}
\newcommand{\mref}[1]{\ref{#1}}

\newcommand{\refFig}[1]{\textrm{Fig.~\mref{#1}}}

\graphicspath{{./graphics/}}
\title{\LARGE \bf
A systems design approach for the co-design of a humanoid robot arm
}

\author{Akhil Sathuluri, Anand Vazhapilli Sureshbabu and Markus Zimmermann
\thanks{The authors are with the Robot Systems Group, Laboratory for Product Development and Lightweight Design, TUM School of Engineering and Design, Technical University of Munich (TUM), Germany
        {\tt\small akhil.sathuluri@tum.de}}%
}

\begin{document}
\maketitle
\thispagestyle{empty}
\pagestyle{empty}

\begin{abstract}
\gist{Context and problem}
Classically, the development of humanoid robots has been sequential and iterative. Such \emph{bottom-up} design procedures rely heavily on intuition and are often biased by the designer's experience. Exploiting the non-linear coupled design space of robots is non-trivial and requires a systematic procedure for exploration. \gist{Methods and ingredients} We adopt the \emph{top-down} design strategy, the \emph{V-model}, used in automotive and aerospace industries. Our co-design approach identifies non-intuitive designs from within the design space and obtains the maximum permissible range of the design variables as a \emph{solution space}, to physically realise the obtained design. \gist{Result} We show that by constructing the solution space, one can (1) decompose higher-level requirements onto sub-system-level requirements with tolerance, alleviating the ``chicken-or-egg" problem during the design process, (2) decouple the robot's morphology from its controller, enabling greater design flexibility, (3) obtain independent sub-system level requirements, reducing the development time by parallelising the development process. 
\end{abstract}

\section{Introduction} \mlabel{sc:intro}
\gist{design problem, State-of-the-art}
Conventional bottom-up design involves several iterations between various sub-systems like mechanical design, actuation and control to ensure compatibility. Further, these sub-systems are tightly coupled, and their non-linear relationship is not intuitive to comprehend and is time-consuming to navigate. Top-down design process provides an alternative to designing such systems by decomposing the higher-level requirements into sub-system-level requirements. One such approach is the combined optimisation of two or more sub-systems as a co-design problem. Studies such as~\mcite{allison2014special, hazard2020automated} have shown that co-design strategy resulted in robots performing better in energy efficiency, time taken, maximum torque required etc., compared to experience-based robot designs. However, the physical feasibility of such robots remains challenging due to issues such as manufacturability, component availability, cost, etc. This highlights the necessity for an alternative design method to identify unique robots and facilitate their physical production.

\section{Methodology}
\begin{figure}[ht]
\includegraphics[width=\linewidth]{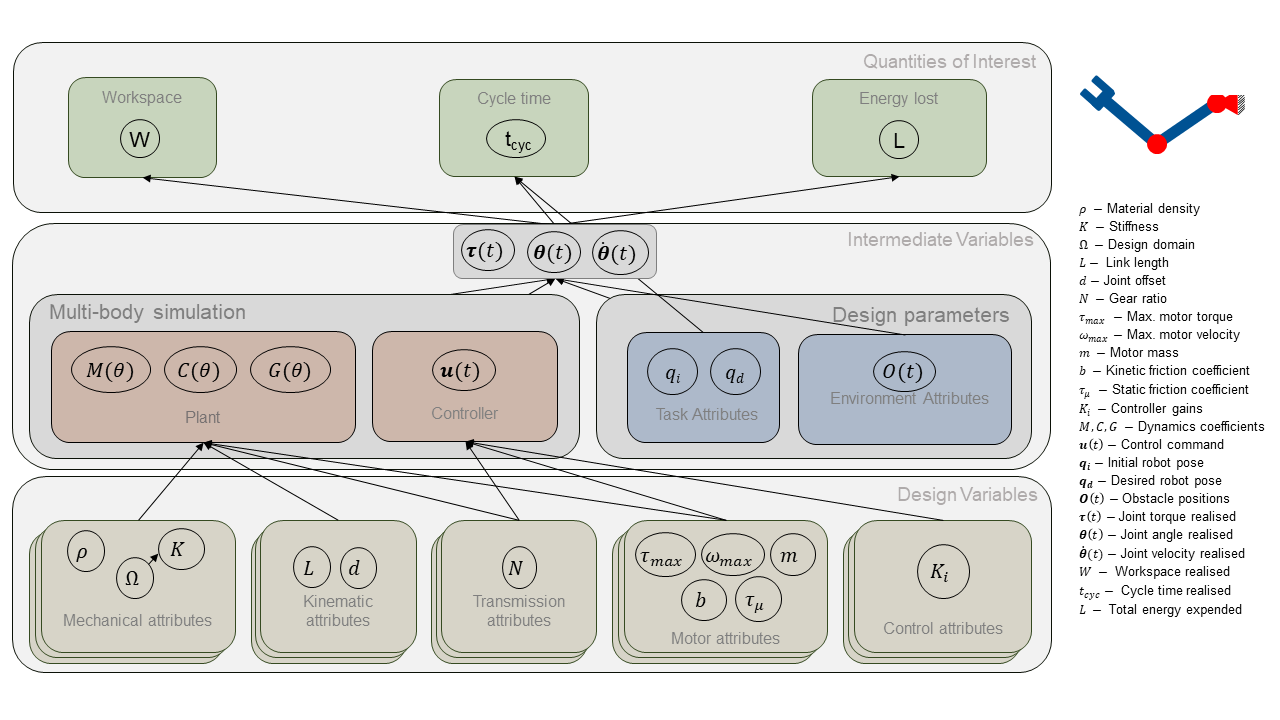}
\centering
\caption{Attribute dependency graph (ADG) mapping the design variables (DVs) to the quantities of interest (QoI)}
\label{fig:adg}
\end{figure}

\gist{solution-space-approach}
We present an optimisation scheme for the co-design of robots based on the V-model~\cite{haskins2006systems}. At the core of the V-model is the definition of user requirements in the initial design phase, which are used as a baseline for design evolution. These system-level goals are then broken down into sub-system-level goals with built-in tolerance by constructing a solution space. Consider a warehouse robot tasked to sort objects between two different bins. Requirements corresponding to the high-level goals are defined for quantities of interest (QoIs), the energy consumed $L\leq6.8e3~J$ and time taken to complete the sorting task, $t_{cyc}\leq3.05~s$. While the required pick and place locations in the workspace, $W$ are enforced as constraints. The relation between the QoIs and the design variables (DVs) is represented as a directed acyclic graph called an attribute dependency graph (ADG)~\cite{zimmermann2017design} as shown in~\refFig{fig:adg}.

The solution space approach involves finding the largest permissible range of the DVs that satisfy the defined requirements. This can be mathematically defined as,
\begin{align}
    \mlabel{eq:obj2}
    \max_{\boldsymbol{\zeta}} &~\mu(\Omega(\boldsymbol{\zeta})),  \\
      \textrm{such that,} \quad & t_{cyc} \leq t_{cyc}(\boldsymbol{x}_{baseline}), \nonumber  \\
    & L \leq L(\boldsymbol{x}_{baseline}) \nonumber 
\end{align}
where $\boldsymbol{\zeta}=(x_{1, l}, x_{1, u}, \dots, x_{d, l}, x_{d, u})$, contains permissible upper and lower bounds of all DVs, $\boldsymbol{x}_1\dots \boldsymbol{x}_d$. 
Further, $\Omega~=~[x_{1, l}, x_{1, u}]\times\dots\times[x_{d, l}, x_{d, u}]$, represents the solution space as a Cartesian product of individual DV ranges. $\mu(\Omega)$ is the size of the corresponding solution space, and $\boldsymbol{x}_{baseline}$ is the baseline design, selected from a reference design or resulting from a conventional co-design optimisation.
\section{Discussion}
The solution space corresponds to the largest hypercube in the design space that satisfies the defined requirements, visualised in 2D as design sections as shown in~\refFig{fig:solutionspace}. The obtained solution space can be employed for:

\begin{figure*}[ht] 
    \centering
    \begin{subfigure}{0.3\textwidth}
        \includegraphics[width=\linewidth,trim={0 0.1cm 0 0},clip]{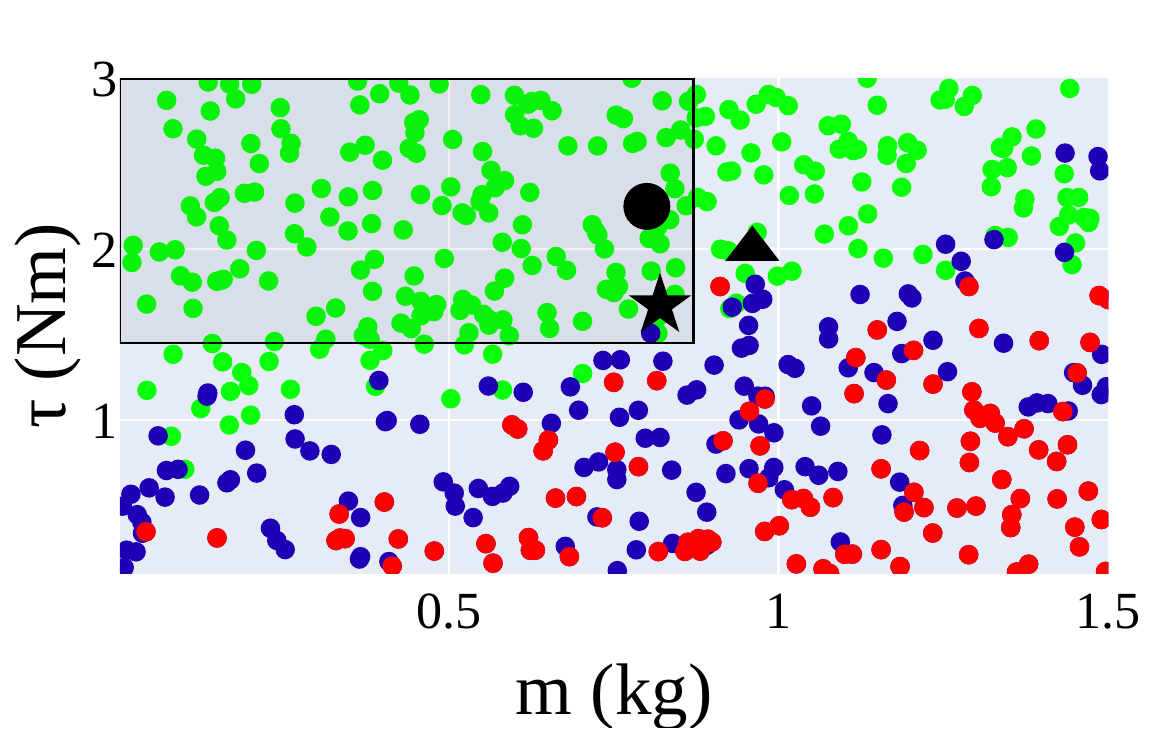}
    \end{subfigure}
    \begin{subfigure}{0.3\textwidth}
        \includegraphics[width=\linewidth,trim={0 0.1cm 0 0},clip]{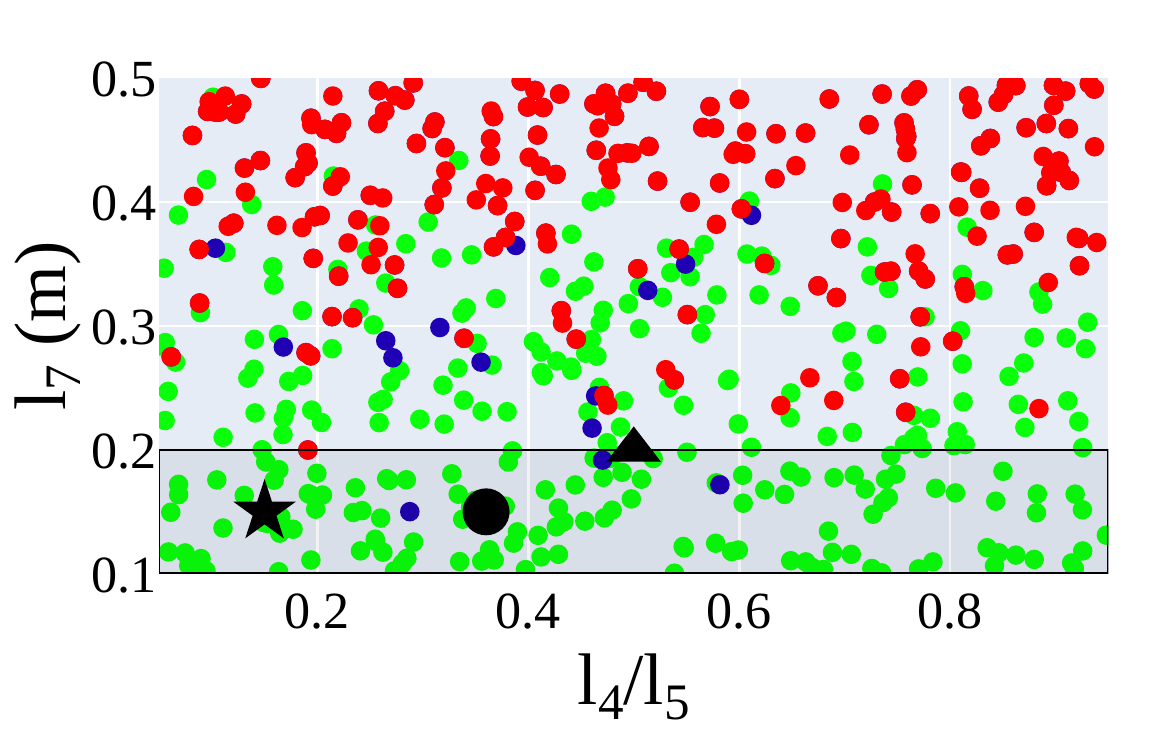}
    \end{subfigure}
    \begin{subfigure}{0.3\textwidth}
        \includegraphics[width=\linewidth,trim={0 0.01cm 0 0},clip]{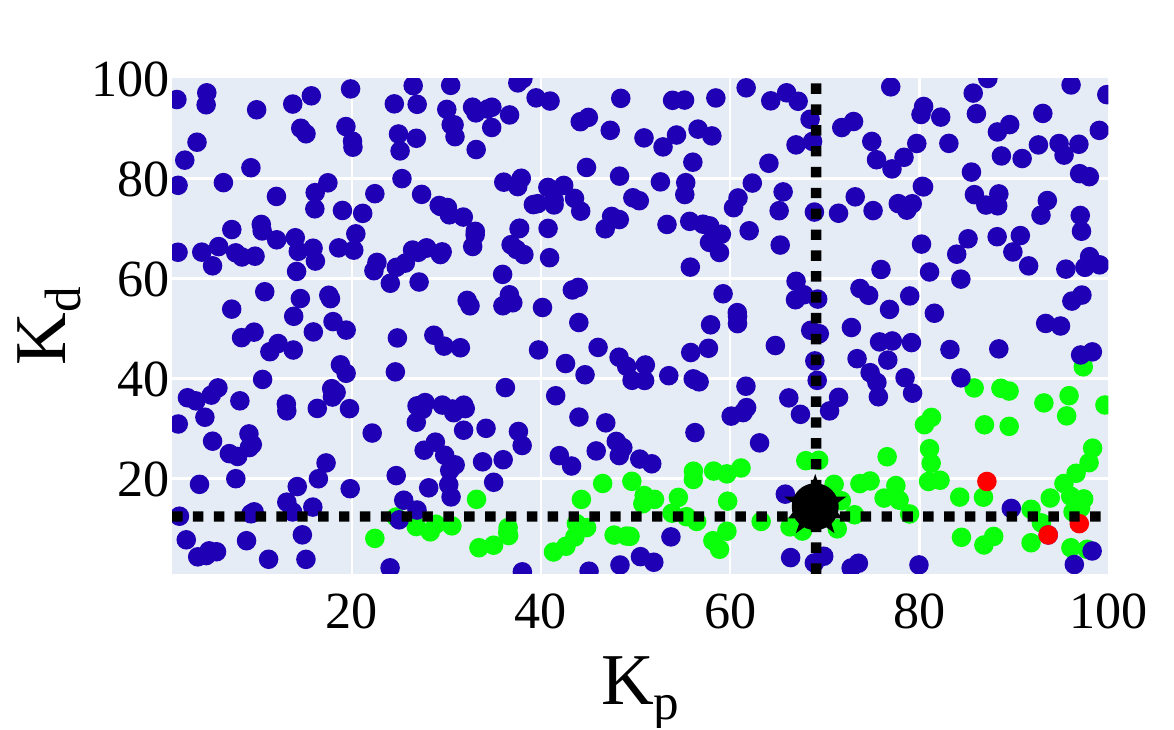}
    \end{subfigure}
    \caption{The rectangular boxes overlaid on each plot are the projections of the solution space. \tikzcircle[green, fill=green]{3pt}, \tikzcircle[red, fill=red]{3pt}, \tikzcircle[blue, fill=blue]{3pt} represent designs that are good and that violate the~$t_{cyc}$ and~$L$ requirements. \tikzcircle[black, fill=black]{3pt}, $\boldsymbol{\blacktriangle}, \boldsymbol{\bigstar}$ are sampled designs of which $\boldsymbol{\bigstar}$ is shown in~\refFig{fig:diva}. DVs $\tau, m, K_p, K_d$ are variables associated only with the first joint of the robot. $l_7$ is the length of link-7, and $l_4/l_5$ is the ratio of two links affecting the motor's location. Note that in each diagram, values for those DVs not shown on the axes are chosen randomly from within the intervals defined by the box edges for other DVs as discussed in~\mcite{zimmermann2017design}
    }   \label{fig:solutionspace}
\end{figure*}

\subsubsection{Sim-to-real for physical feasibility}
Generally, simulation engines are leveraged to find an optimal design iteratively. However, for computational tractability, simplified robot models are utilised. Unlike traditional optimisation approaches that result in point-based solutions, solution space provides acceptable ranges of all the DVs, allowing one to select components such as motors or gearboxes from a larger space. Moreover, the ranges act as tolerances resulting in improving robustness. As a result, the divergence from the simplified simulation models is readily accommodated by solution spaces.
\\
\subsubsection{Parallelising the design process}
The solution space provides permissible ranges of the DVs realised without affecting the system-level performance. For example, a motor DV value can be selected from the first plot of~\refFig{fig:solutionspace} independent of the length of the links realised from the second plot as long as they are within the specified bounds, parallelising the development of different sub-systems. 
\\
\subsubsection{Decoupling of mechanics and control}
Similarly, the solution sub-space in the mechanics DVs represents the largest variations in the robot's morphology independent of the choice of the control DVs and vice versa. Solution spaces provide this unique ability to decouple the robot's mechanics and control, allowing for greater design flexibility.  
\\
\subsubsection{Design space trade-off}
Control variables are purely logical and can be modified easily. Their design space can be traded off to obtain larger ranges for other DVs as shown in~\refFig{fig:solutionspace}. This is made possible by the coupled design space of the co-design problem in~\refFig{fig:adg}. The design range for one DV can be translated into the corresponding design range of any other DV via the solution space. 
\\
\subsubsection{Interpretation of the design space}
The design sections in~\refFig{fig:solutionspace} illustrate the local boundary between good and bad designs. These boundaries provide insights into the interdependence between their corresponding DVs. The X-Ray toolbox\footnote{https://github.com/akhilsathuluri/x-ray-tool} provides an interface for the designer to trade-off and interpret the resulting non-linear design space. 

\begin{figure}[t]
\includegraphics[width=0.75\linewidth, trim={3cm 0 5cm 0}, clip]{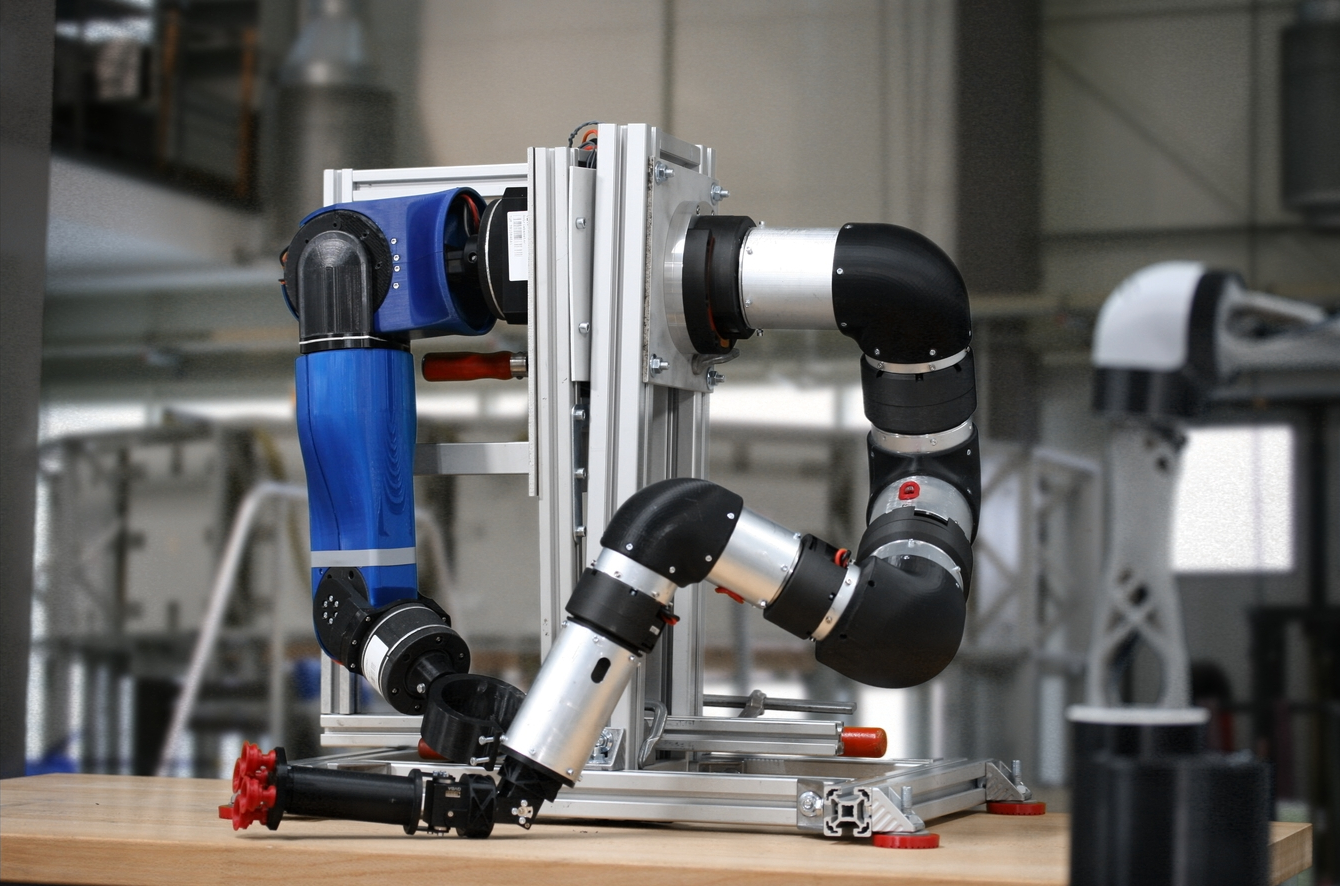}
\centering
\caption{Left: A robot with classical humanoid arm kinematics developed bottom-up, Right: Robot resulting from a top-down development procedure}
\label{fig:diva}
\end{figure}

Figure~\ref{fig:diva} illustrates the differences in the design processes, where for the top-down design, 
requirements were obtained from a point-based co-design optimisation, followed by the solution space approach. The video\footnote{https://youtu.be/QydGXRZJV60} shows the robot performing the task.


\bibliographystyle{IEEEtran}
\bibliography{stewart}

\end{document}